%% file: main.tex
\documentclass{article}
\usepackage{arxiv}

\usepackage[utf8]{inputenc} 
\usepackage[T1]{fontenc}    
\usepackage{hyperref}       
\usepackage{url}            
\usepackage{booktabs}       
\usepackage{amsfonts}       
\usepackage{nicefrac}       
\usepackage{microtype}      
\usepackage{lipsum}		
\usepackage{graphicx}
\usepackage{natbib}
\usepackage{doi}
\usepackage{url}
\usepackage{foot
note}
\usepackage{tablefootnote} 
\makesavenoteenv{tabular}

\usepackage[usenames,dvipsnames]{xcolor}
\usepackage{array}
\usepackage{CJKutf8}
\usepackage{balance}
\usepackage{multirow}
\usepackage{subcaption}
\usepackage{placeins}
\graphicspath{ {figs/} }
\usepackage{amsmath, amssymb}
\usepackage{mathtools} 
\usepackage{autobreak}
\usepackage{type1cm}
\usepackage{environ}
\NewEnviron{myEquation}{
\begin{equation}
\scalebox{0.9}{$\BODY$}
\end{equation}
}
\usepackage{xspace}
\usepackage{bm}

\newcommand{\eg}{\textit{e.g.}}

\newcommand{\methodname}{\textit{\textbf{CLanG}\xspace}}

\title{Multi-object event graph representation learning for Video Question Answering}

\author{Yanan Wang \quad Shuichiro Haruta\quad Donghuo Zeng \quad Julio Vizcarra \quad Mori Kurokawa \vspace{0.3em} \\
{KDDI Research} \\
{{\tt \small\{wa-yanan,do-zeng,sh-haruta,xdo-zen,xju-vizcarra,mo-kurokawa\}@kddi.com} \quad
}
}

\date{}

\renewcommand{\headeright}{}

\begin{document}
\maketitle

\input{000_abstract}
\input{001_introduction}
\input{002_relatedwork}
\input{003_architecture}

\input{004_experiment}

\input{005_conclusion}

\bibliographystyle{unsrtnat}
\bibliography{references}  






\end{document}

%% file: 000_abstract.tex
\begin{abstract}
Video question answering (VideoQA) is a task to predict the correct answer to questions posed about a given video. The system must comprehend spatial and temporal relationships among objects extracted from videos to perform causal and temporal reasoning. While prior works have focused on modeling individual object movements using transformer-based methods, they falter when capturing complex scenarios involving multiple objects (\eg, ``a boy is throwing a ball in a hoop"). 
We propose a contrastive language event graph representation learning method called~\methodname~to address this limitation.
Aiming to capture event representations associated with multiple objects, 
our method employs a multi-layer GNN-cluster module for adversarial graph representation learning, enabling contrastive learning between the question text and its relevant multi-object event graph. 
Our method outperforms a strong baseline, achieving up to \textbf{2.2\%} higher accuracy on two challenging VideoQA datasets, NExT-QA and TGIF-QA-R.
In particular, it is \textbf{2.8\%} better than baselines in handling causal and temporal questions, highlighting its strength in reasoning multiple object-based events. 
\end{abstract}

%% file: 001_introduction.tex
\section{Introduction}

\begin{figure}[t]
\centering
\includegraphics[width=0.6\columnwidth]{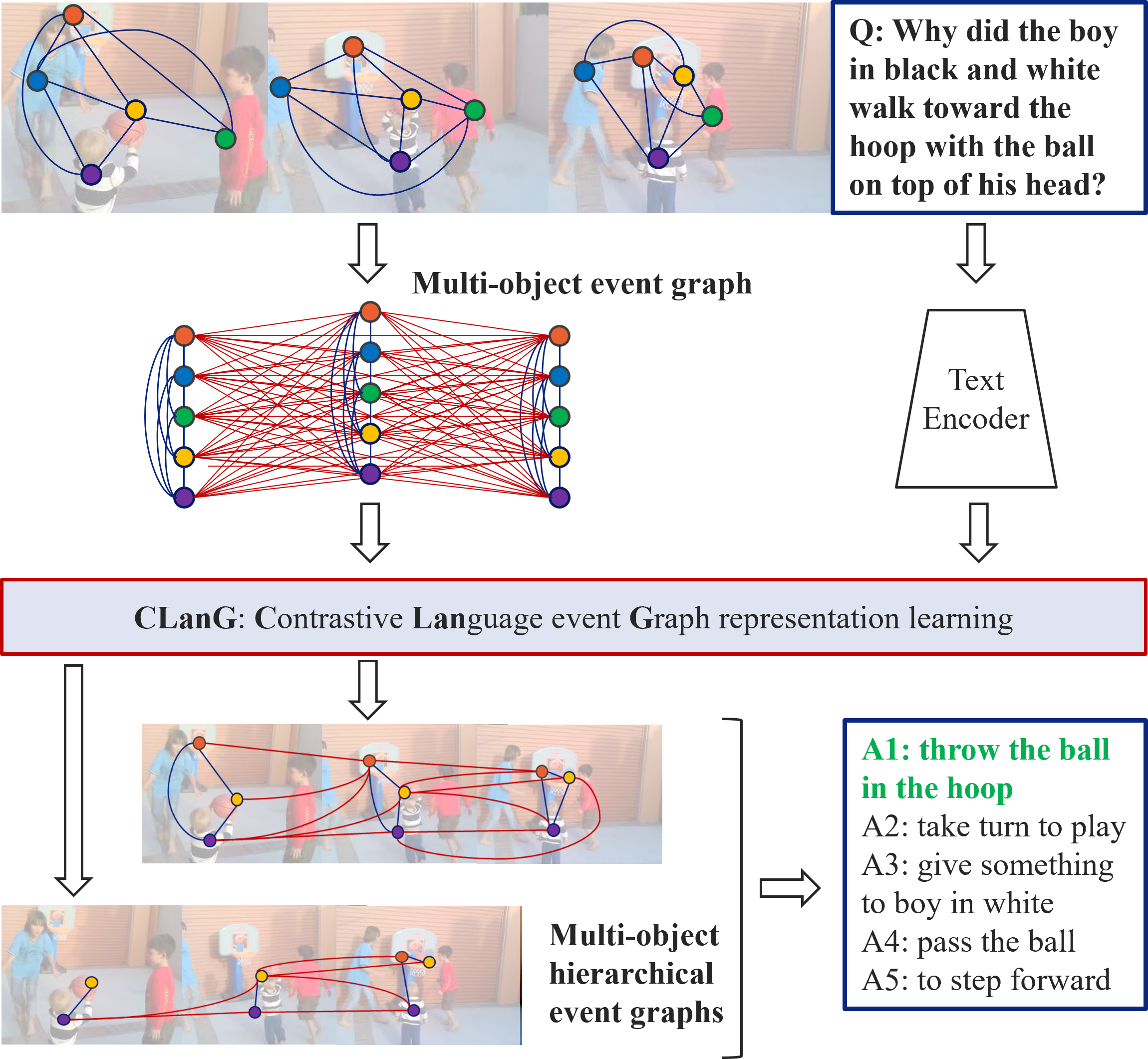}
\caption{Overview of \textit{\methodname}. Our proposed method obtains multi-object hierarchical event graph representations for causal and temporal reasoning in videos by giving a fully connected multi-object event graph.}
\label{fig:CLanG}
\end{figure}

Video question answering (VideoQA) is a task to predict the correct answer to questions posed about a given video. It extends the concept of question answering from static images (\eg, VQA~\cite{antol2015vqa,wang2023vqagnn}) to dynamic videos. 
Recent works~\cite{yang2021justask, Lei_2021_CVPR, gao2023mist} have fine-tuned visual-language pretraining models(VLMs) to acquire holistic video representations for frame-level question answering. 
To enhance causal and temporal reasoning in VideoQA~\cite{xiao2021next}, recent works~\cite{jin2021adaptive, Geng2021DynamicGR, xiao2022video} employ graph neural networks (GNNs) for capturing precise object-level event details.
These approaches model single-object events by tracking their motions, such as ``a boy is walking", and ``a ball is moving".
However, they fall short in capturing multi-object events like ``a boy is throwing a ball into a hoop," which involve three objects (the ``boy", ``ball", and ``hoop").
This limitation arises due to the absence of temporal relationships between multiple objects, hindering the system's ability to capture multi-object event representations for enhancing causal and temporal question answering in videos.


This work proposes a contrastive language event graph representation learning method called~\methodname~to capture hierarchical event representations associated with multiple objects (see Fig.~\ref{fig:CLanG}). 
It contains a multi-layer GNN-cluster module and two training objectives for effective event graph representation learning.
Inspired by recent graph pooling approaches~\cite{ying2018hierarchical,bo2020sdcn,bianchi2020mincutpool,liu2022graph} that coarsening large real graphs such as a social network graph into smaller sizes for extracting useful clusters, the multi-layer GNN-cluster module expands a graph pooling method~\cite{ying2018hierarchical} to multiple layers to obtain multi-scale event graphs.
In addition, to eliminate the loss of original graph information with the proposed multi-layer GNN-cluster module, we reduce the size of the output graphs during the first half of the GNN-cluster layers. Conversely, in the latter half, we increase the number of output nodes to match the input graph.
The multi-layer GNN-cluster module also involves a self-attention layer to highlight the impact of hierarchical event graphs.
We further perform adversarial graph representation learning~\cite{pan2018adversarially} with the output of the last GNN-cluster layer to enforce all node representations to follow regular distributions to improve the language event graph contrastive learning process.

We end-to-end train \methodname~with the target VideoQA task on three challenging datasets, NExT-QA~\cite{xiao2021next}, and TGIF-QA-R~\cite{peng_2021_MM}. By comparing with strong baselines, \methodname~achieves up to \textbf{2.2\%} accuracy improvement on NExT-QA and TGIF-QA-R. In particular, it is \textbf{2.8\%} better than baselines for temporal questions, suggesting its strength in performing multiple object-based event question reasoning. 

%% file: 002_relatedwork.tex
\section{Related work}

\textbf{Spatio-temporal graph-based VideoQA methods.} Recent works~\cite{L-GCN2020AAAI, Geng2021DynamicGR, xiao2021video, wang2023vqagnn} unify graph neural networks (GNN) with a transformer to capture the relationship between visual objects for achieving fine-grained object-level scene understanding. The work~\cite{xiao2022video} adds temporal edges to track individual objects across the built scene graph of each video frame. However, it focuses on capturing the single-object event by giving a well-structured spatio-temporal graph. In contrast, our model captures multi-object event representations with a multi-layer GNN-cluster module by giving a fully connected multi-object event graph across the whole video. 

\textbf{Graph pooling methods.} Graph pooling is an essential element of GNN architecture for obtaining a holistic graph-level representation of the entire graph~\cite{liu2022graph}. Recent works~\cite{ying2018hierarchical, bo2020sdcn, bianchi2020mincutpool} propose a differentiable graph pooling module to learn how to generate hierarchical representations of graphs, have broadly adapted to graph classification and graph generation tasks. However, the capability of these models is always limited by the low number of layers. Our method enables multi-layer graph poolings by computing adversarial graph representation learning and language-graph contrastive learning for capturing multi-scale hierarchical event representations.

%% file: 003_architecture.tex
\section{Methodology}\label{sec:method}

\begin{figure*}[t]
\centering
\includegraphics[width=0.95\linewidth]{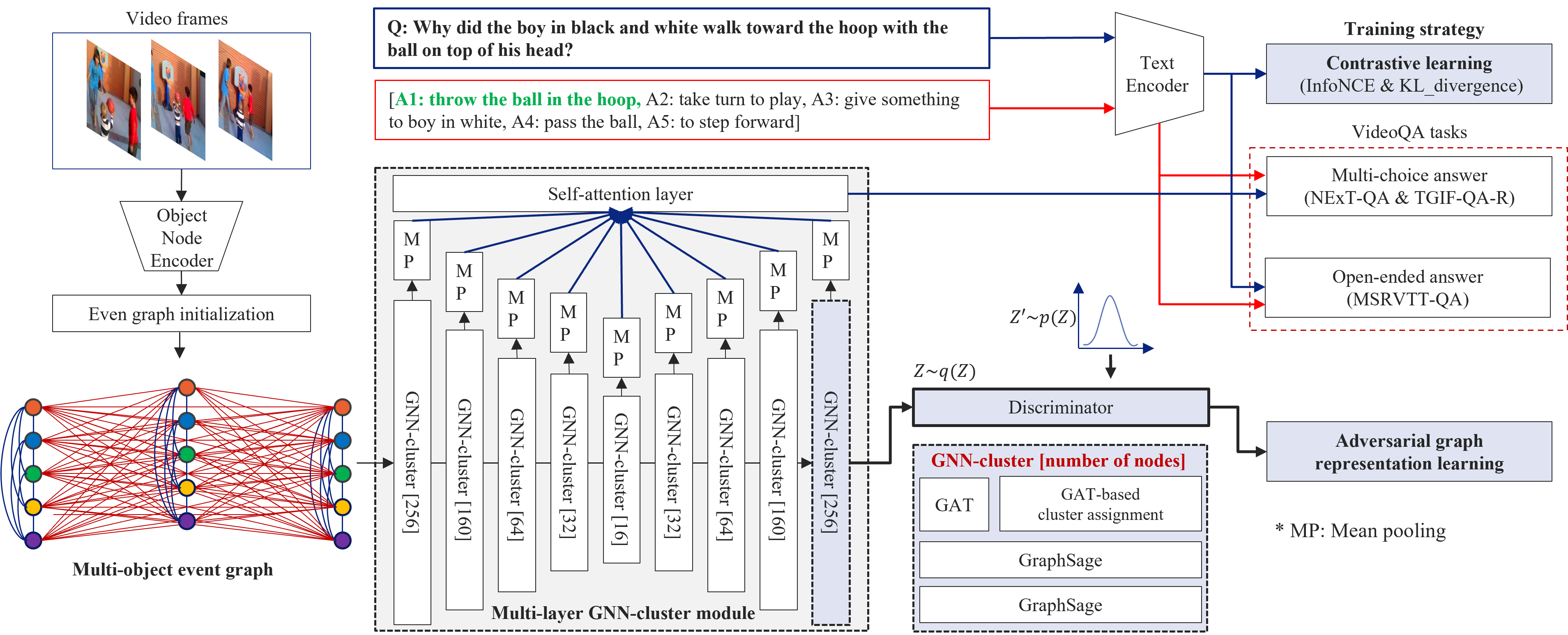}
\captionsetup{width=\textwidth}\caption{\methodname~architecture. We first initialize a dense adjacency matrix with encoded object nodes to build a \textbf{multi-object event graph}, then we apply a \textbf{multi-layer GNN-cluster module} with a discriminator for \textbf{adversarial graph representation learning}, where the multi-layer GNN-cluster module consists of nine GNN-cluster involving the different number of object nodes. Finally, We perform \textbf{language event graph contrastive learning} and video question answering with encoded text and multi-object event graph representations.}
\label{fig:clang_architecture}
\end{figure*}


\subsection{Multi-object event graph Initialization} \label{sec:eventgraph}
We segment a video into $K$ clips, with each clip containing $L$ frames, following the way in~\cite{xiao2021video}. We then extract $N$ RoI features $X_{r}^{N\times d1}=\{x_{r_n}^{d1}\}_{n=1}^{N}$ along with their corresponding spatial position $X_{b}^{N\times d2}=\{x_{b_n}^{d2}\}_{n=1}^{N}$ from each frame. This extraction is accomplished through a pretrained object detector~\cite{anderson2018bottom}. We initialize object node representations as follows:
\begin{eqnarray}\label{eq:0}
   & X^{N\times (d1+d2)}=\{(x_{r_n}^{d1}||x_{b_n}^{d2})\}_{n=1}^{N}
\end{eqnarray}
where the ``$||$" symbol indicates the concatenation operation.

To construct a multi-object event graph from a given video, we define the graph nodes $X=\{x_m\}_{m=1}^M~(M=K\times L\times N)$, involving $N$ extracted objects across $L$ frames and $K$ clips. 
We also define a spatial relation score $sr(i,j)$ to quantify the similarity between distinct nodes within the same frame:
\begin{eqnarray}\label{eq:1}
   & sr(i,j) = \psi(x_i^t,x_{j}^t|j \in \mathcal{N}_i)
\end{eqnarray}
where $\psi$ denotes the cosine similarity between two detected objects $x_i^t$ and $x_j^t$ within the same frame $t$, while $\mathcal{N}_i$ represents all other nodes except node $i$.

Unlike the approach in prior work~\cite{xiao2022video}, which aligns the same object across adjacent frames, we introduce a temporal relation score $tr(i,j)$ to quantify the similarity between distinct nodes across different video frames:
\begin{eqnarray}\label{eq:2}
   & tr(i,j) = \psi(x_i^t,f_{j \in \mathcal{N}_i}^{\bar{t}}); ~(t,~\bar{t}) \in \{0,1,..., K\times L\}
\end{eqnarray}
where $t$ and $\bar{t}$ indicate any different frames in a video. We unify the spatial and temporal relation scores into a common matrix called spatio-temporal adjacency matrix $A=\{a_{ij}\}_{i=1,j=1}^M$, where $a_{ij} \in \{sr(i,j), tr(i,j)\}$. 
The multi-object event graph is defined as $\mathcal{G}=(A, X)$.

\subsection{Multi-layer GNN-cluster module}\label{sec:gnncluster}
The Multi-layer GNN-cluster module is a critical component of our approach, aimed at modeling expressive hierarchical event representations using the built multi-object event graph (see Fig.~\ref{fig:clang_architecture}).
This module comprises an $N$-layer GNN-cluster, which serves to coarsen the input graph $(A^{(l)}, X^{(l)})$ into a new coarsened graph $(A^{(l+1)}, X^{(l+1)})$, while also learning a cluster assignment matrix $S^{(l)} \in \mathbf{R}^{n_l\times n_{l+1}}$.
This assignment matrix learning follows the DIFFPOOL architecture~\cite{ying2018hierarchical}.
To enhance the assignment matrix learning, we integrate two GAT~\cite{velickovic2017graph} modules into the original DIFFPOOL design. These GAT modules enable better capturing of mutual information between neighboring nodes.
The GNN-cluster computation is performed by inputting the graph $\mathcal{G}=(A^{(l)}, X^{(l)})$ using the following steps:
\begin{eqnarray}\label{eq:3}
   & Z^{(l)} = \operatorname{GAT}_{l,embed}(A^{(l)},X^{(l)}), \\
   & S^{(l)} = \operatorname{softmax}(\operatorname{GAT}_{l,pool}(A^{(l)},X^{(l)})),\\
   & X^{(l+1)} = S^{(l)^T}Z^{(l)} \in \mathbf{R}^{n_{l+1}\times d},\\
   & A^{(l+1)} = S^{(l)^T}A^{(l)}S^{(l)} \in \mathbf{R}^{n_{l+1}\times n_{l+1}}
\end{eqnarray}
where $Z^{(l)}$ represents the output of the embedding GAT module. Additionally, $n_{l}$ denotes the number of input nodes, and $n_{l+1}$ denotes the number of nodes in the output graph. The GNN-Cluster further incorporates two GraphSage layers~\cite{hamilton2017inductive}, which follow the GAT modules and compute event graph representations at the $(l+1)$-th layer.

\methodname~utilizes a multi-step approach involving $N$-layer GNN-cluster followed by mean pooling layers to generate multi-scale event graph representations. 
Our approach begins by downsizing the output graphs in the initial GNN-cluster layers, effectively coarsening the input graph. Conversely, in the subsequent GNN-cluster layers, we expand the output node count to match that of the input graph, thereby retaining the original graph information.
Finally, the resulting multi-scale pooled graph representations are input into a self-attention layer, yielding a latent event graph representation $X_g$ to dynamically emphasize key hierarchical event graph features.

\subsection{Adversarial graph representation learning}\label{sec:advglearning}
To enhance the training of the multi-layer GNN-cluster module, we introduce a discriminator $D$ to enforce the latent node representations $Z$ (the output of the final GNN-cluster $\mathcal{G}^{(l+1)} = (A^{(l+1)}, X^{(l+1)})$) to match a normal distribution $p_z$.
The discriminator $D$ constructed as a standard multi-layer perception (MLP), generates a one-dimensional output unit followed by a sigmoid layer.
We compute the following cross-entropy cost $\mathcal{L}^{D}$ to distinguish whether the input of $D$ is from $p_z$ (positive) or from the GNN-cluster (negative).
\begin{equation}\label{eq:4}
\mathcal{L}^{D} = -\mathbb{E}_{\mathbf{z}\sim p_z}\log D(\mathbf{z})-\mathbb{E}_{\mathbf{X^{(l+1)}}}\log (1-D(\mathcal{G}^{(l+1)})
\end{equation}
Meanwhile, we compute a regularization loss termed $\mathcal{L}^{\mathcal{G}^{(l+1)}}$ for the GNN-cluster to ensure the regularization of node representations.
\begin{equation}\label{eq:5}
\mathcal{L}^{\mathcal{G}^{(l+1)}} = -0.5\times \mathbb{E}_{\mathbf{X^{(l+1)}}}\log (D(\mathcal{G}^{(l+1)})
\end{equation}
The adversarial graph representation learning is performed by minimizing both $\mathcal{L}^{D}$ and $\mathcal{L}^{\mathcal{G}^{(l+1)}}$. 

\subsection{Language event graph contrastive learning}\label{sec:contlearning}
Utilizing BERT LM, We encode the question text to obtain the question embedding $X_q$. Given a batch of $N$ ($X_q, X_g$) pairs, \methodname~is trained to predict the positive pair to learn a joint embedding space across language and visual modalities. This process enables the multi-layer GNN-cluster module of the model to learn textual event knowledge within the event graph nodes. We apply contrastive learning by minimizing InfoNCE~\cite{oord2018representation} loss, which is formulated as follows:
\begin{equation}\label{eq:6}
\mathcal{L}_{N}=-\log{\frac{e^{\mathrm{sim}(X_q, X_g^+)/\tau}}{ e^{\mathrm{sim}(X_q, X_g^+)/\tau } + \sum_{i=1}^{K}{e^{\mathrm{sim}(X_q, X_{i}^-)/\tau }}}}
\end{equation}
where $\mathrm{sim}(X_q, X_g^+)$ denotes the cosine similarity of the positive pair. $(X_q, X_{i}^-)$ denotes the negative pair and $K$ is the number of negative pairs in a batch, set to $N^2-N$. The temperature $\tau$ is set to $0.1$. 

In addition, we minimize Kullback-Leibler divergence loss $\mathcal{L}_{KL}$ for the positive pairs to match the probability distribution of the text encoder and the multi-layer GNN-cluster module~\cite{pkt_eccv}. $\mathcal{L}_{KL}$ is formulated as:
\begin{eqnarray}\label{eq:7}
&\mathcal{P}_{q} = \frac{\mathrm{sim}(X_q,X_q^T)}{\sum_{i=1}^{N}{\mathrm{sim}(X_q,X_q^T)}_i},
\mathcal{P}_{g} = \frac{\mathrm{sim}(X_g,X_g^T)}{\sum_{i=1}^{N}{\mathrm{sim}(X_g,X_g^T)}_i},\\
&\mathcal{L}_{KL} = \frac{1}{N}{\sum_{i=1}^{N}(\mathcal{P}_{g[i]} \log{\frac{\mathcal{P}_{q[i]}}{\mathcal{P}_{g[i]}}}})
\end{eqnarray}


\subsection{Video question answering}
\methodname~reasons the correct answer $a$ by leveraging the encoded event graph representations $X_g$ and question embedding $X_q$.
For the multi-choice QA task (\eg, NExTQA and TGIF-QA-R), multiple choices $\mathcal{A}_{mc}$ are provided along with each question. All candidates $a\in \mathcal{A}_{mc}$ are encoded using the same encoder employed for the question text.
In contrast, for the open-ended QA task, each question corresponds to a broader set of candidates $\mathcal{A}_{o}$, where only one candidate $a \in \mathcal{A}_{o}$ is the most suitable answer. 

During the training process, \methodname~is end-to-end optimized by minimizing a unified loss $\mathcal{L}$ given by:
\begin{equation}\label{eq:7}
\mathcal{L} =\mathcal{L}_{D} +\mathcal{L}^{\mathcal{G}^{(l+1)}}+ \mathcal{L}_{N}+\mathcal{L}_{KL}+\mathcal{L}_{QA}
\end{equation}
where $\mathcal{L}_{QA}$ denotes the softmax cross entropy loss for the downstream video question answering task.
The adversarial graph representation learning ($\mathcal{L}_{D} +\mathcal{L}^{\mathcal{G}^{(l+1)}}$) and language event graph contrastive learning ($\mathcal{L}_{N}+\mathcal{L}_{KL}$) are incorporated to facilitate the extraction of multi-object hierarchical event representations.

%% file: 004_experiment.tex
\section{Experiments}\label{sec:experiment}

\subsection{Dataset}
We evaluate \methodname~on three challenging videoQA datasets (Tab.~\ref{tab:dataset}), NExT-QA~\cite{xiao2021next}, TGIF-QA-R~\cite{peng_2021_MM}. 
NExT-QA is a multi-choice videoQA benchmark featuring causal and temporal questions involving object-level spatio-temporal reasoning.
TGIF-QA-R is a reconstruction of TGIF-QA~\cite{jang-IJCV-2019}, wherein all ground-truth answers are collected as part of an answer vocabulary. Candidate answers are then randomly selected from this vocabulary to mitigate answer biases. This reconstruction amplifies the challenge, particularly in the question type of state transition. 

\begin{table}[h]

\centering
\begin{tabular}{l|l|c|c|c|c|c}
\toprule
Datasets & \#QAs & Train & Val & Test & VLen (s) & QA \\
\midrule
NExT-QA & 48K & 34K & 5K & 9K & 44 & MC \\
TGIF-QA-R & 58.9K & 52.7K & - & 6.2K & 3 & MC \\
\midrule
\end{tabular}
\caption{Dataset Details. ``MC" and ``OE" denote multi-choice and open-end VideoQA benchmarks, respectively.}
\label{tab:dataset}
\end{table}

\subsection{Training details}
We decode the video into $K=4$ clips, and each clip contains $L=8$ frames. We utilize top $N=20$ object nodes for NExT-QA and $N=10$ for TGIF-QA-R to achieve our best scores. The dimension of \methodname’s hidden states is $d = 512$. The model is optimized with AdamW under a learning rate of $1$e-$5$. The batch size is $32$ and the models are trained for $30$ epochs on a single NVIDIA Quadro RTX 8000 GPU.

\subsection{Performance}
In Tab.~\ref{tab:1}, CLanG-BERT achieves \textbf{2.2\%} higher averaging test accuracy for all question types compared to a strong baseline method VGT~\cite{xiao2022video}. In particular, our method scores \textbf{2.8\%} better at handling causal and temporal questions since it can comprehend multiple events connected to various objects.
VGT focuses on tracking \textit{single}-object temporal relationships within individual video clips to derive object-level spatio-temporal event representations, it encounters challenges in modeling complex, long-form multi-events involving multiple objects.
In contrast, \methodname~constructs a \textit{multi}-object event graph utilizing a Multi-layer GNN-cluster module across the entire video sequence, enabling the acquisition of multi-object hierarchical event representations. 
Moreover, CLanG-Roberta achieves a validation accuracy of \textbf{59.15\%} for causal reasoning and \textbf{60.55\%} for all question types. Even though MIST-CLIP~\cite{gao2023mist} adapts a visual-language pretraining model for intricate long-form videoQA, CLanG-Roberta surpasses MIST-CLIP by \textbf{3.3\%} in the realm of causal reasoning, suggesting its efficacy in executing language event graph representation learning.
This, in turn, empowers our novel Multi-layer GNN-cluster module to extract multi-object event representations from the RoBERTa-based model. 
Moreover, a comparison between CLanG-BERT and CLanG-RoBERTa indicates the capacity of a potent large-scale language Model (LLM) to elevate performance through language event graph contrastive learning.
Tab.~\ref{tab:2} shows the comparison results on TGIF-QA-R dataset. CLanG-RoBERTa improves VGT on the state transition types by over \textbf{1.9\%} test accuracy, suggesting the efficacy of temporal reasoning. 

\begin{table*}[ht!]

\centering
\resizebox{\linewidth}{!}{%
\begin{tabular}{l|cccc|cccc}  
\toprule
\multirow{2}{*}{Model} & \multicolumn{4}{c}{Val. Acc.(\%)} & \multicolumn{4}{c}{Test Acc.(\%)} \\ 
\cmidrule(lr){2-9}

& Causal & Temporal & Descriptive & All & Causal & Temporal & Descriptive & All \\
\midrule
VQA-T
&41.66 &44.11 &59.97 &45.30 &42.05 &42.75 &55.87 &44.54\\
HQGA
&48.48 &51.24 &61.65 &51.42 &49.04 &52.28 &59.43 &51.75\\
VGT
&52.28 &55.09 &64.09 &55.02 &51.62 &51.94 &63.65 &53.68\\
MIST-CLIP
&54.62 &56.64 &66.92 &57.18 &- &- &- &-\\
\midrule
CLanG-BERT &53.24($\pm 0.02$) &54.78($\pm 0.02$) &64.48($\pm 0.03$) &55.48($\pm 0.02$) &53.73($\pm 0.01$) &54.69($\pm 0.02$)&65.05($\pm 0.03$) &55.89($\pm 0.02$)\\
\textbf{CLanG-RoBERTa} &\textbf{59.15}($\pm 0.02$) &\textbf{58.81}($\pm 0.01$) &68.98($\pm 0.05$) &\textbf{60.55}($\pm 0.02$) &\textbf{60.11}($\pm 0.01$) &\textbf{60.03}($\pm 0.02$) &\textbf{66.98}($\pm 0.03$) &\textbf{61.21}($\pm 0.02$) \\
\bottomrule
\end{tabular}
}
\captionsetup{width=\textwidth}\caption{Comparison results with strong baselines on NExT-QA.}
\label{tab:1}
\end{table*}




\begin{table}[h]

\centering
\begin{tabular}{l|c|c|c}  
\toprule
Model & GNN-Cluster num. & Object num. & Test Acc.(\%)\\ 
\midrule
HCRN &- & 10 & 63.9\\
PGAT &- & 10 & 65.9\\
VGT &- & 10 & 70.5\\
\midrule
CLanG-BERT&8 & \multirow{2}{*}{10} & 71.0\\
CLanG-RoBERTa&12 & & \textbf{72.4}\\

\bottomrule

\end{tabular}
 \caption{Accuracy scores of state transition question type for TGIF-QA-R.}
\label{tab:2}
\end{table}




\subsection{Ablation study}
\textbf{Effect of Multi-layer GNN-cluster.} The results in Tab.~\ref{tab:4} demonstrate that the $8$-layer Multi-layer GNN-cluster improves the model without any GNN-cluster layer by \textbf{1.1\%} in test accuracy for NExT-QA. While \methodname~without GNN-cluster still manages to enhance the prior methods due to the proposed language event graph representations learning, it fails to capture multi-scale hierarchical event graphs essential for further enhancing causal and temporal reasoning.

\begin{table}[h]

\centering
\begin{tabular}{l|c|c}  
\toprule
Model & Numbers of GNN-clusters &Test Acc.(\%) \\
\midrule
\multirow{3}{*}{CLanG-BERT} & 0 &54.71 \\
& \textbf{8} &\textbf{55.89} \\
& 12 &54.16 \\
\midrule
CLanG-RoBERTa & 12 &\textbf{61.21} \\
\midrule
\end{tabular}
\caption{Effect of the Multi-layer GNN-cluster module on NExT-QA.} 
\label{tab:4}
\end{table}

\textbf{Effect of training strategies.} To study the effect of the proposed adversarial graph representation learning and language event graph representation learning, 
we conducted a comparison between models: CLanG-BERT without adversarial graph representation learning and CLanG-BERT without language event graph representation learning, against the original model (as shown in Tab.~\ref{tab:5}). 
The exclusion of either of these training strategies led to a performance reduction of the original model to \textbf{2.5\%},
suggesting their effectiveness in training expressive hierarchical event representations through the proposed Multi-layer GNN cluster module.

\begin{table}[h]

\centering
\begin{tabular}{l|c}  
\toprule
Model & Test Acc.(\%) \\
\midrule
CLanG-BERT & \textbf{55.89} \\
- w/o adversarial graph representation learning & 54.88 \\
- w/o language event graph contrastive learning  & 53.32 \\
\midrule
\end{tabular}
 \caption{Effect of training strategies on NExT-QA.}
\label{tab:5}
\end{table}

%% file: 005_conclusion.tex
\section{Conclusion} \label{conclusions}
We proposed a novel video question answering method \methodname~that models hierarchical event representations associated with multiple objects for excelling causal and temporal reasoning. 
In the next step, we will expand our work to build a language-graph fundamental model that leads to fine-grained video scene understanding.